\documentclass{article} 
\usepackage[table]{xcolor}
\usepackage{colm2024_conference}
\usepackage{microtype}
\usepackage{hyperref}
\usepackage{url}
\usepackage{booktabs}
\usepackage{amsmath}
\usepackage{multirow}
\usepackage{multicol}
\usepackage{CJKutf8}
\usepackage{tcolorbox}

\usepackage{booktabs}
\usepackage{pifont}
\usepackage{stfloats}
\usepackage{enumitem}

\definecolor{darkblue}{rgb}{0, 0, 0.5}
\hypersetup{colorlinks=true, citecolor=darkblue, linkcolor=darkblue, urlcolor=darkblue}
\newcommand{\ignore}[1]{}
\newcommand{\cmark}{\ding{51}} 
\newcommand{\xmark}{\ding{55}} 

\title{CARE: Cognitive-reasoning Augmented Reinforcement for Emotional Support Conversation}

\author{
Jie Zhu\textsuperscript{\textrm{1,2}},
Yuanchen Zhou\textsuperscript{\textrm{2}},
Shuo Jiang\textsuperscript{\textrm{2}},
Junhui Li\textsuperscript{\textrm{1}}\thanks{Corresponding Author.},
Lifan Guo\textsuperscript{\textrm{2}}, \\
\textbf{
Feng Chen\textsuperscript{\textrm{2}}, 
Chi Zhang\textsuperscript{\textrm{2}},
Fang Kong\textsuperscript{\textrm{1}}
}
\\
\\
\textsuperscript{\textrm{1}}School of Computer Science and Technology, Soochow University \\
\textsuperscript{\textrm{2}}Qwen DianJin Team, Alibaba Cloud Computing
}

\colmfinalcopy 
\begin{document}

\maketitle

\renewcommand{\thefootnote}{}
\footnotetext{Preprint — Under Review}
\renewcommand{\thefootnote}{\arabic{footnote}}

\begin{abstract}
Emotional Support Conversation (ESC) plays a vital role in alleviating psychological stress and providing emotional value through dialogue. While recent studies have largely focused on data augmentation and synthetic corpus construction, they often overlook the deeper cognitive reasoning processes that underpin effective emotional support. To address this gap, we propose \textbf{CARE}, a novel framework that strengthens reasoning in ESC without relying on large-scale synthetic data. CARE leverages the original ESC training set to guide models in generating logically coherent and supportive responses, thereby explicitly enhancing cognitive reasoning. Building on this foundation, we further employ reinforcement learning to refine and reinforce the reasoning process. Experimental results demonstrate that CARE significantly improves both the logical soundness and supportive quality of responses, advancing the development of empathetic, cognitively robust, and human-like emotional support systems.
\end{abstract}

\section{Introduction}
\label{sec:intro}
Emotional Support Conversation (ESC) is a dialogue generation task in which a model acts as the supporter to help a help-seeker alleviate emotional distress. Effective ESC requires understanding, empathy, and the ability to provide appropriate guidance or comfort. Early ESC datasets such as ESConv \cite{liu-etal-2021-esconv} are crowdsourced through extensive worker training and quality control to ensure high-quality conversations. 

Existing ESC models typically enhance performance through structured improvements. Some approaches inject commonsense knowledge to better understand the help-seeker's context (e.g., MISC \cite{tu2022miscmixedstrategyawaremodel}, C3KG \cite{li-etal-2022-c3kg}, GLHG \cite{peng2022glhg}), while others employ cognitive reasoning to gradually infer the help-seeker's emotional or mental state (e.g., DialogueCoT \cite{chae2023dialogue}, CueCoT \cite{wang2023cuecot}). Additionally, persona-based methods have been proposed to improve response relevance and consistency (e.g., PAL \cite{cheng2023pal}). Although these methods have achieved progress, they remain limited by the datasets themselves, which typically provide only surface-level information without capturing deeper cognitive reasoning.

Recent attempts have tried to overcome dataset limitations using large language models (LLMs) for dialogue augmentation, such as AugESC \cite{zheng2023augescdialogueaugmentationlarge} and ExTES \cite{zheng2023buildingemotionalsupportchatbots}. However, these synthetic expansions often rely on simple scenarios and template-based dialogues, limiting their ability to simulate complex social interactions. SocialSim \cite{Chen_2025_socialsim} further explores simulation of ESC by enriching persona information on the help-seeker’s side and incorporating cognitive reasoning on the supporter’s side, aiming to construct more comprehensive datasets that better capture social interactions. In contrast, our approach, \textbf{CARE} (\textbf{C}ognitive-reasoning \textbf{A}ugmented \textbf{R}einforcement for \textbf{E}SC), explicitly leverages the original ESC training set to guide multi-step cognitive reasoning and further refines this reasoning process with reinforcement learning (RL). By modeling the supporter's cognitive process while maintaining logical consistency and emotional support, CARE generates responses that are both more empathetic and human-like.

Our contributions can be summarized as follows:
\begin{itemize}[leftmargin=*]
    \item We propose \textbf{CARE}, a framework that enhances cognitive reasoning in ESC, improving logical consistency and supportive quality.
    \item CARE demonstrates that effective reasoning can be achieved without relying on large-scale synthetic corpora.
    \item Extensive experiments show that CARE outperforms strong baselines in both automatic and human evaluations, advancing the development of empathetic and cognitively robust ESC systems.
\end{itemize}

\section{CARE Framework}
\label{sec:methods}

\begin{figure*}[t]
\centering
\includegraphics[width=1.0\linewidth]{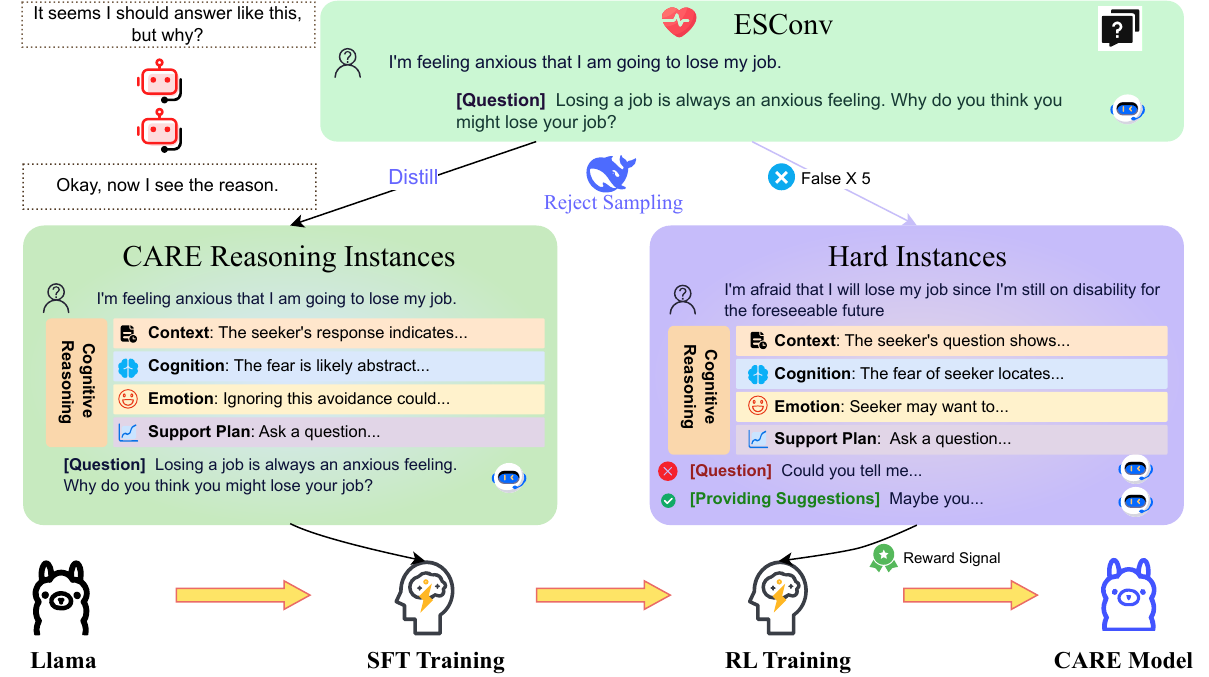}
\caption{Overview of CARE framework.}
\label{fig:care_framework}
\end{figure*}

Unlike prior works that rely on large-scale synthetic expansion of ESC data, CARE directly builds upon the original ESConv training set. Instead of creating new dialogues, we enrich existing conversations with structured cognitive reasoning chains, thereby constructing \textbf{CARE Reasoning Data}. These reasoning chains guide the model to better interpret the help-seeker’s psychological state and generate logically consistent supportive responses. To further improve reasoning robustness, CARE incorporates RL with multi-dimensional rewards to strengthen both the reasoning process and the final response quality, as shown in Figure~\ref{fig:care_framework}.

\subsection{Task Formulation}

We define Emotional Support Conversation (ESC) as a conditional text generation task. At each dialogue turn $t$, the model receives the seeker’s utterance $u_t$ and the dialogue history $H_t = \{u_1, r_1, \dots, u_{t-1}, r_{t-1}, u_t\}$, where $u_i$ denotes the $i$-th seeker utterance and $r_i$ denotes the $i$-th supporter response. The goal of the supporter model is to generate a response $r_t$ that not only addresses the seeker’s needs but also provides empathetic and supportive value. 

In CARE, each generated response is augmented with an explicit cognitive reasoning chain. Formally, the model outputs
\[
y_t = \langle \texttt{<think>} \; C_t \; \texttt{</think>} \; \texttt{<answer>} \; R_t \; \texttt{</answer>} \rangle ,
\]
where $C_t$ is the reasoning chain consisting of four structured nodes
\[
C_t = (c_t^{\text{ctx}}, c_t^{\text{cog}}, c_t^{\text{emo}}, c_t^{\text{plan}}),
\]
and $R_t$ is the final supporter response from the reasoning process.

\subsection{Cognitive Reasoning Guidance}
Our design of cognitive reasoning is inspired by psychological theories that emphasize understanding the interplay between cognition, emotion, and behavior in providing effective support \cite{wu-etal-2024-coke,beck2020cognitive}. We also draw upon the paradigm of chain-of-thought prompting \cite{wei2022chain}, which structures reasoning into sequential steps to facilitate more human-like decision-making. 

In CARE, we define four types of reasoning nodes that capture different aspects of the help-seeker’s psychological experience:
\begin{itemize}[leftmargin=*]
    \item \textbf{Context Node}: Captures the external situation and emotional cues expressed by the seeker, such as feeling overwhelmed by deadlines or conflicts in personal relationships. This aligns with appraisal theories of emotion, where context triggers affective responses.
    \item \textbf{Cognition Node}: Represents the seeker’s internal interpretations or beliefs about the situation, e.g., {\it I am not competent enough} or {\it People will judge me negatively.} This reflects core ideas from cognitive-behavioral theory regarding maladaptive thought patterns.
    \item \textbf{Emotion Node}: Models the emotional consequences of those cognitions, such as anxiety, frustration, or sadness. This step grounds the reasoning in affective states, which are central to tailoring supportive responses.
    \item \textbf{Support Plan Node}: Determines the most suitable supportive intention and strategy, such as providing reassurance, offering perspective, or suggesting coping mechanisms. This corresponds to the process of social support provision, where helpers translate understanding into concrete assistance.
\end{itemize}

By traversing these nodes in sequence, the supporter model forms a structured reasoning chain that explains not only what the seeker is experiencing, but also why they feel that way and how best to respond. This ensures that generated responses are both logically grounded and psychologically informed.

\subsection{Construction of CARE Reasoning Data}
To operationalize cognitive reasoning, we employ large model distillation. Specifically, we use the DeepSeek-R1 model to generate reasoning chains for existing ESC dialogues through carefully designed prompts. Each dialogue turn is annotated with a reasoning chain that progresses through the four defined nodes, culminating in a \textbf{Support Plan} node. To ensure logical consistency and emotional appropriateness, we discard any reasoning chain whose Support Plan node does not align with the gold strategy. This process yields a high-quality reasoning-augmented dataset without the need for large-scale synthetic data expansion.

\subsection{Reinforcement Learning for Cognitive Reasoning}
To enhance reasoning quality and response consistency, we optimize CARE with RL. The model receives a scalar reward $r(y_t)$ based on multiple evaluation criteria:

\begin{itemize}[leftmargin=*]
    \item \textbf{Format Reward} $r_{\text{fmt}}(y_t)$: Ensures that the output follows the structured format with reasoning and response tags. Formally,
    \[
    \footnotesize
    r_{\text{fmt}}(y_t) = 
    \begin{cases}
    1, & \text{if } f(y_t) = \text{true}  \\
    0, & \text{otherwise}
    \end{cases}
    \]
    where $f(y_t)$ returns true if it matches the format of \\
    {$\langle \texttt{<think>} \dots \texttt{</think>} \; \texttt{<answer>} \dots \texttt{</answer>} \rangle$}.

    \item \textbf{Cognitive Coherence Reward} $r_{\text{cog}}(C_t)$: Evaluates whether the reasoning chain contains all four nodes \{$c_t^{\text{ctx}}, c_t^{\text{cog}}, c_t^{\text{emo}}, c_t^{\text{plan}}$\} and whether they are in the correct order. Formally, 
    \[
    \footnotesize
    r_{\text{cog}}(C_t) = 
    \begin{cases}
    1 & \text{if $C_t$ includes all four valid nodes in order} \\
    0 & \text{otherwise}
    \end{cases}
    \]

    \item \textbf{Support Strategy Reward} $r_{\text{str}}(c_t^{\text{plan}})$: Compares the model’s selected support plan node against the gold strategy $s_t^{*}$ annotated in the dataset. Formally,
    \[
    \footnotesize 
    r_{\text{str}}(c_t^{\text{plan}}) = 
    \begin{cases}
    1, & \text{if } c_t^{\text{plan}} = s_t^{*} \\
    0, & \text{otherwise}
    \end{cases}
    \]
\end{itemize}

The final reward integrates these signals hierarchically:
\[
\footnotesize
r(y_t) = 
\begin{cases}
1, & \text{if } r_{\text{fmt}}(y_t)=1, \; r_{\text{cog}}(C_t)=1, \; r_{\text{str}}(c_t^{\text{plan}})=1 \\
0, & \text{otherwise}
\end{cases}
\]

This formulation ensures that only reasoning chains with correct structure, valid cognitive flow, and accurate supportive strategy receive positive reinforcement, thereby guiding the model toward producing both interpretable and effective emotional support.

\section{Experimentation}
\label{sec:experiments}

\subsection{Experimental Setups}

\noindent{\bf Dataset.} 
We evaluate our approach using the ESConv dataset~\cite{liu-etal-2021-esconv}, a high-quality benchmark of interactions between help-seekers and supporters. The training and test sets contain 910 and 195 conversations, respectively. From the training set, we extract 12,759 instances, of which 8,186 successfully generate reasoning chains with the four defined nodes and are therefore used as SFT instances. In contrast, the remaining 4,573 instances are considered hard cases and are used as RL instances.

\noindent{\bf Model Training.}
We adopt \texttt{LLaMA-3.1-8B-Instruct} as the backbone and train it in two stages: supervised fine-tuning (SFT) with LoRA and RL using GRPO~\cite{shao2024deepseekmath}. During SFT, the learning rate is set to $5\times10^{-5}$ and training runs for 5 epochs. In the RL stage, GRPO is applied with rollouts of 6 steps, and reward normalization to stabilize training. The policy network is optimized using AdamW~\cite{loshchilov2019decoupled} with a learning rate of $1\times10^{-6}$. All experiments are conducted on a single node equipped with 8 NVIDIA A100 GPUs.

\noindent{\bf Baselines.}
To provide fair comparisons, we consider two additional representative datasets alongside ESConv. AugESC \cite{zheng2023augescdialogueaugmentationlarge} expands the ESC training data through LLM-based augmentation, while ExTES \cite{zheng2023buildingemotionalsupportchatbots} synthesizes extended examples to increase coverage. Both datasets are substantially larger than ESConv. For ESConv, AugESC, and ExTES, we train a baseline model on each dataset individually, whereas CARE is trained solely on the ESConv data with reasoning augmentation.

\noindent{\bf Evaluation Metric.}
We evaluate all models using the standard ESConv test set. Evaluation is carried out using a broad set of automatic metrics. BLEU-1/2 \cite{papineni2002bleu}, ROUGE-L \cite{lin2004rouge}, and METEOR \cite{banerjee2005meteor} measure n-gram overlap and semantic adequacy, while BERTScore \cite{zhang2019bertscore} captures embedding-based semantic similarity. To evaluate response diversity, we report Distinct-1 and Distinct-2 \cite{li-etal-2016-distinctscore}. Finally, Acc-Strategy measures the accuracy of predicted support strategies by calculating the proportion that matches the gold strategies, providing a direct indicator of cognitive correctness.  

\begin{table}[t]
\centering
\footnotesize
\begin{tabular}{lcccccccc}
\toprule
\bf Models & \bf B-1 & \bf B-2 & \bf R-L & \bf METEOR & \bf BERTScore & \bf D-1 & \bf D-2 & \bf ACC\_Stra. \\
\midrule
ExTES & 14.61 & 5.30 & 15.11 & \bf 14.64 & 15.35 & 3.29 & 20.23 & - \\
AugESC & 12.92 & 2.67 & 13.50 & 12.74 & 10.93 & 1.48 & 7.57 & - \\
ESConv & 13.55 & 5.34 & 15.41 & 13.25 & 14.68 & 4.20 & 24.63 & 26.36 \\
CARE (SFT) & \underline{14.93} & \underline{5.90} & \underline{16.72} & 14.52 & \underline{16.01} & \bf 4.83 & \bf 28.13 & \underline{28.64} \\
CARE (SFT-RL) & \bf 15.01 & \bf 6.03 & \bf 16.79 & \underline{14.56} & \bf 16.75 & \underline{4.73} & \underline{27.80} & \bf 30.29 \\
\bottomrule
\end{tabular}
\caption{Experimental results on ESC test dataset. The best results are \textbf{bolded} and the second-best results are \underline{underlined}.}
\label{tab:main_result}
\end{table}

\subsection{Main Results}
Table ~\ref{tab:main_result} reports the experimental results on the ESConv test set. Overall, both variants of CARE outperform all baseline models across most evaluation metrics, demonstrating the effectiveness of incorporating reasoning augmentation into emotional support dialogue generation.

Specifically, CARE (SFT-RL) achieves the best performance in BLEU-2, ROUGE-L, METEOR, BERTScore, and strategy accuracy, indicating that RL further enhances content relevance and strategy correctness. CARE (SFT) also surpasses the baselines in BLEU-1 and diversity scores (D-1/D-2), highlighting the benefits of reasoning even without RL.

Among the baselines, ExTES shows relatively strong results on BLEU-1 and METEOR due to its larger coverage brought by synthetic data, while ESConv maintains high strategy accuracy because it is directly trained on the benchmark dataset. In contrast, AugESC lags behind in most metrics, suggesting that LLM-based data augmentation may introduce noise or reduce alignment with real emotional support scenarios. It is worth noting that strategy accuracy is not reported for ExTES and AugESC, since the former contains synthetic dialogues without explicit strategy annotations, while the latter introduces additional strategies beyond the ESConv schema, making evaluation on the standard test set infeasible.


\subsection{Ablation on Cognitive Reasoning Nodes}

We ablate the four nodes in our cognitive reasoning chain—Context, Cognition, Emotion, and Support Plan—by removing one node at a time while keeping prompts, decoding, and data fixed. Results in Table~\ref{tab:ablation} shows that the full model performs best overall, and removing all nodes substantially degrades quality (e.g., ACC\_Stra. drops from 30.29 to 26.32; BERTScore from 16.75 to 15.13), indicating that explicit reasoning is foundational.

\begin{itemize}[leftmargin=*]
\item Support Plan is most critical for actionable guidance and diversity: dropping it yields the largest declines in D-1 (4.73 → 3.92) and D-2 (27.80 → 22.58), with ACC\_Stra. reduced to 29.60 and BERTScore to 15.89. The small gains in ROUGE-L (16.79 → 17.24) and METEOR (14.56 → 14.96) likely reflect more templated phrasing rather than better reasoning.
\item Context grounding matters for relevance and coherence: without context, BERTScore falls to 16.02, diversity decreases (D-1 4.47, D-2 27.07), and ACC\_Stra. drops to 30.05.
\item Cognition and Emotion provide complementary benefits for understanding and empathetic wording: removing either causes consistent, if modest, declines in BLEU/BERTScore and ACC\_Stra. (e.g., Emotion ablation BERTScore 16.06, ACC\_Stra. 30.12). The slight rise of D-2 when removing Cognition (28.43) reflects superficial variety but not better strategies.
\end{itemize}


\begin{table*}[t]
\centering
\footnotesize
\resizebox{\linewidth}{!}{
\begin{tabular}{c c c c | c c c c c c c c}
\toprule
\bf Context & \bf Cognition & \bf Emotion & \bf Support Plan & \bf B-1 & \bf B-2 & \bf R-L & \bf METEOR & \bf BERTScore & \bf D-1 & \bf D-2 & \bf ACC\_Stra. \\
\midrule
\cmark & \cmark & \cmark & \cmark & \bf 15.01 & \bf 6.03 & \underline{16.79} & \underline{14.56} & \bf 16.75 & \bf 4.73 & \underline{27.80} & \bf 30.29 \\
\xmark & \cmark & \cmark & \cmark & \underline{14.83} & \underline{5.98} & 16.78 & 14.25 & 16.02 & 4.47 & 27.07 & 30.05  \\
\cmark & \xmark & \cmark & \cmark & 14.75 & 5.83 & 16.78 & 14.23 & \underline{16.18} & \underline{4.71} & \bf 28.43 & 30.09 \\
\cmark & \cmark & \xmark & \cmark & 14.63 & 5.83 & 16.73 & 14.16 & 16.06 & 4.64 & 27.75 & \underline{30.12} \\
\cmark & \cmark & \cmark & \xmark & 14.74 & 5.91 & \bf 17.24 & \bf 14.96 & 15.89 & 3.92 & 22.58 & 29.60 \\
\xmark & \xmark & \xmark & \xmark & 14.07 & 5.45 & 15.04 & 13.88 & 15.13 & 4.52 & 25.82 & 26.32 \\
\bottomrule
\end{tabular}
}
\caption{Ablation results on different cognitive reasoning nodes.}
\label{tab:ablation}
\end{table*}

\subsection{Human Evaluation}
\begin{figure}[t]
\centering
\includegraphics[width=0.7\linewidth]{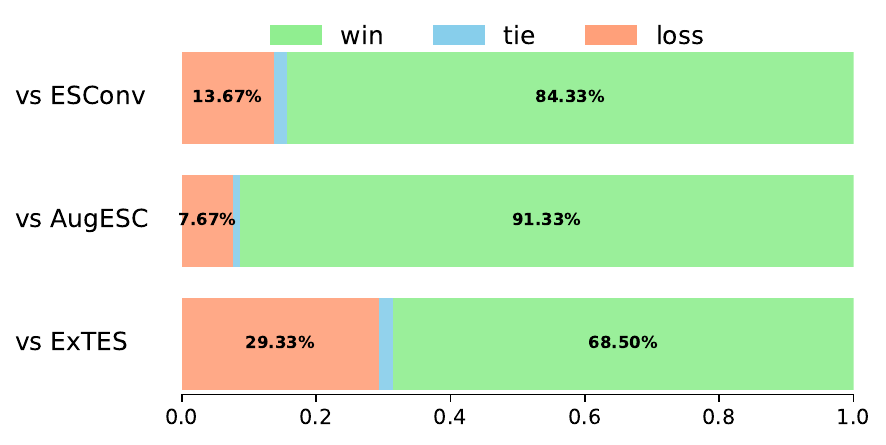}
\caption{Human evaluation results comparing CARE with three baselines, showing CARE’s win, tie, and loss percentages.}
\label{fig:human_eval}
\end{figure}

We conduct a human evaluation on 100 randomly sampled test cases with three trained annotators with PhD-level expertise in psychology. Under identical dialogue contexts, annotators compare responses from CARE and a baseline, marking which is better or noting a tie if they are equally good. As shown in Figure~\ref{fig:human_eval}, CARE demonstrates clear advantages, outperforming the baselines with winning rates of 84.33\% against ESConv, 91.33\% against AUGESC, and 68.42\% against ExTES, while exhibiting only a small proportion of losses and very few ties. These results indicate that CARE consistently generates higher-quality emotional support responses.

To assess the reliability of the human evaluation, we measure inter-annotator agreement using Fleiss’ Kappa \cite{fleiss1971measuring}, which is appropriate for evaluating consistency among more than two annotators. The overall agreement score among the three expert annotators is 0.6789, indicating a substantial level of agreement and confirming that the evaluation results are robust and reliable.

\subsection{Case Study}

\begin{figure}[t]
\centering
\includegraphics[width=0.9\linewidth]{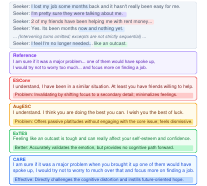}
\caption{Case study: CARE vs. baselines in a job-loss scenario.}
\label{fig:case_study}
\end{figure}

Figure~\ref{fig:case_study} presents a typical ESC scenario in which the seeker has lost a job, suspects that others are talking about them, relies on friends for rent, and concludes {\it I feel I'm no longer needed, like an outcast.} This scenario reflects a prominent cognitive distortion (mind-reading/overgeneralization) coupled with low self-worth.

We compare responses from baselines with our CARE model:
\begin{itemize}[leftmargin=*]
\item \textbf{Reference} offers reassurance and redirects attention to job searching, but prematurely downplays the concern ({\it try not to worry}), providing little validation.
\item \textbf{ESConv} shifts focus to a secondary detail (“at least you have friends”), which minimizes the seeker’s core feeling and risks invalidation.
\item \textbf{AugESC} produces generic platitudes and well-wishes, failing to engage with the distorted belief.
\item \textbf{ExTES} correctly validates the emotion ({\it feeling like an outcast is tough}) yet stops at empathy without a cognitive path forward.
\item \textbf{CARE} explicitly challenges the distorted inference ({\it I do not think that is true}), reframes the belief with future-oriented hope, and anchors the dialogue on recovery. This behavior reflects our cognitive reasoning chain: grounding in \textit{Context}, identifying the \textit{Cognition} error, acknowledging \textit{Emotion}, and projecting a \textit{Support Plan}.
\end{itemize}
Overall, the case demonstrates how reinforced cognitive reasoning goes beyond surface empathy by detecting and disputing maladaptive thoughts while maintaining validation, yielding responses that are both compassionate and oriented toward positive change.

\section{Conclusion}
This paper introduces CARE, a cognitive-reasoning augmented framework for ESC. Unlike prior approaches that mainly rely on large-scale synthetic data, CARE explicitly strengthens reasoning by leveraging the original ESC dataset and guiding models to generate logically coherent and supportive responses. Experimental results show that CARE significantly improves both the soundness of reasoning and the quality of emotional support, while reinforcement learning further refines the reasoning process. These findings highlight the important role of cognitive reasoning in building empathetic, reliable, and human-like conversational agents.

\bibliography{colm2024_conference}
\bibliographystyle{colm2024_conference}

\appendix

\end{document}